    \documentclass[handcarry]{aiaa-tc}
    \usepackage{amsmath}
    \usepackage{verbatim}
    \usepackage{graphicx}
    \usepackage{xcolor}
    \graphicspath{{images/}}
    \usepackage{subcaption}
    \usepackage{epstopdf}
    \usepackage{gensymb}
    \usepackage{tabularx}
    \usepackage{booktabs} 
    \usepackage{multirow}
    \usepackage{arydshln} 
    \usepackage{url}
    \usepackage{hyperref} 
    \usepackage{textcomp} 
    \usepackage[titletoc,title]{appendix}
    \usepackage{float} 
    \usepackage{color}
    \usepackage{setspace}
    \usepackage{fancyhdr}
    \title{Training Detection-Range-Frugal Cooperative Collision Avoidance Models for Quadcopters via Neuroevolution}

    \author{
    	Amir Behjat
    	\thanks{Ph.D. Student, Department of Mechanical and Aerospace Engineering. Email: amirbehj@buffalo.edu}\thanksibid{1}\\
    	{\normalsize\itshape
    		University at Buffalo, Buffalo, NY, 14260}
    	\and
    	Krushang Gabani
    	\thanks{MS Student, Department of Mechanical and Aerospace Engineering. Email: krushang@buffalo.edu}\\
    	{\normalsize\itshape
    		University at Buffalo, Buffalo, NY, 14260}\\
    	\and
    	Souma Chowdhury
    	\thanks{Assistant Professor, Department of Mechanical and Aerospace Engineering. Email: soumacho@buffalo.edu}\\
    	{\normalsize\itshape
    		University at Buffalo, Buffalo, NY, 14260}
    }

    \providecommand{\keywords}[1]{\textbf{\textit{keywords: }} #1}
    
\begin{document}
    	\maketitle	
    	\begin{abstract}
    Cooperative autonomous approaches to avoiding collisions among small Unmanned Aerial Vehicles (UAVs) is central to safe integration of UAVs within the civilian airspace. One potential online cooperative approach is the concept of reciprocal actions, where both UAVs take pre-trained mutually coherent actions that do not require active online coordination (thereby avoiding the computational burden and risk associated with it). This paper presents a learning based approach to train such reciprocal maneuvers. Neuroevolution, which uses evolutionary algorithms to simultaneously optimize the topology and weights of neural networks, is used as the learning method -- which operates over a set of sample approach scenarios. Unlike most existing work (that minimize travel distance, energy or risk), the training process here focuses on the objective of minimizing the required detection range; this has important practical implications w.r.t. alleviating the dependency on sophisticated sensing and their reliability under various environments. A specialized design of experiments and line search is used to identify the minimum detection range for each sample scenarios. In order to allow an efficient training process, a classifier is used to discard actions (without simulating them) where the controller would fail. The model obtained via neuroevolution is observed to generalize well to (i.e., successful collision avoidance over) unseen approach scenarios. 
    	
    	
    \vspace{0.5cm}	
    \noindent\keywords{Autonomous, Cooperative Collision Avoidance, Detection Range,  Neuroevolution, Unmanned Aerial Vehicles (UAVs)}
    		
    \end{abstract}

    	\section{Introduction}\label{secintro}

    Successful introduction of small low-altitude and low-speed unmanned aerial vehicles (UAVs) into the airspace is dependent on the maturity of technologies associated with their operational safety. Avoiding collisions with tall infrastructure, and other small UAVs and manned aircraft sharing the airspace is central to these technologies. It is also important to note that increasing market growth will cause more UAVs to operate in close proximity to each other, thereby increasing the likelihood of UAV-UAV encounters. Thus, unsurprisingly, there is a growing body of work in UAV collision avoidance, mostly in the context of static obstacles or non-cooperative UAV-UAV encounters. However, there is limited understanding of whether mutually reciprocal behavior, e.g., flexible traffic rules, could play a crucial role in guiding the online maneuver decisions in the case of friendly encounters (most commercial/civilian operations arguably falls in this category). To address this knowledge gap, this paper focuses on developing and testing a cooperative collision avoidance scheme for multirotor UAVs, wherein both UAVs implement the same collision avoidance model, learned offline. The primary contribution of the paper lies in presenting a novel (collision-avoidance) maneuver design approach that exploits a learning paradigm called neuroevolution to train generalizable neural network-based action models. Herein the learning process optimizes over an objective that measures the range of detection required to allow robust avoidance, with the goal to reduce dependency on sophisticated detect/sense capability assumptions. The remainder of this section briefly reviews the literature on UAV collision avoidance and converges on the objectives of the research presented here.
    
    \subsection{Survey of Collision Avoidance Methods}
    
    Existing collision avoidance methods can be broadly classified \cite{all} into: geometric, optimized trajectory, bearing angle, force field and Markov decision process approaches.
    
    Geometric Approach alters the trajectory of the UAVs based on the pose of the ownship and intruder UAV \cite{kumar}. When a collision is detected \cite{EKFcollision}), each UAV is given a priority number, and the one with lower priority chooses a way-point that is perpendicular to its velocity. While this method can seek energy-optimal paths, it is not guaranteed that changing the path is the optimal strategy in the first place. Graph search approaches include A* and Dijkstra's algorithms \cite{A*}, which have been extended from typically dealing with stationary obstacles, to UAV-UAV collision scenarios \cite{grad_desc,gan2012real,richards2002aircraft} with limited applicability.

    Some of the other online maneuver planning approaches tend to be computationally expensive. For instance, the Bearing Angle Approach, which uses cameras to estimate the relative angle of the obstacle concerning the UAV and collision is prevented by keeping the obstacle image at a desired safe position in the camera's field of view \cite{angb}\cite{Yang2013}. Then, there are Force-field approaches, which use attraction/repulsion principles from classical robotics. In some implementations, proximity to another UAV incites a braking force \cite{1272619}, while in other \cite{shim2003decentralized}, the current states of all UAVs are used to predict the trajectory over a time horizon. 
    Another approach for collision avoidance between UAV is using Markov decision process (MDP) and Partially Observable MDP (POMDP) \cite{pomdp,pomdp2} methods. 
    While POMDPs are potent in providing system-aware optimal and safe actions under uncertainties, the online computing burden can become intractable, and performance is highly sensitive to tunable parameters \cite{pomdp}.  
    
    \subsection{Research Objectives: Reciprocal Collision Avoidance}
    
    In this paper, we build upon our new bio-inspired mutually reciprocal collision avoidance method, targeted at autonomous quadcopter UAV operations. This method uses two different collision avoidance strategies where each UAV can temporarily change its heading (thereby deviating from their original path) or change its speed to avoid a collision. Figure \ref{fig:Col} shows the schematic path of a UAV when \textit{speed change} or \textit{direction change} maneuvers are applied. As can be seen from Fig. \ref{fig:Col}, the action to be decided by the maneuver model (a neural network system) is encoded in terms of the amount of change in speed or heading angle, and the time of maneuver initiation w.r.t. the collision detection time point. These actions are to be taken in response to the estimated pose of the other UAV, where the relative state of the two UAVs can be completely represented with a total of five degrees of freedom; these parameters serve as inputs to the maneuver decision model. The two main quantities of interest that drive the training of the optimum action model are 1) the minimum separation distance experienced during the maneuver (enforced as a constraint) and 2) the detection range required to allow a just-feasible maneuver w.r.t. the separation constraint.

    Autonomous UAVs need a collision detection and sensing system to detect potential collisions with static and dynamic obstacles. 
    These collision detection systems use information such as object position, orientation, velocity, size, and maneuverability \cite{detect_01}, extracted from sensors such as stereo vision and LIDAR. The detection range of the sensing system may vary according to weather conditions, velocity, and orientation of the UAVs. Given the complexity and computational cost \cite{detect_02} associated with detection and pose estimation of the other system and its dependence on external factors, it is desirable to teach UAVs avoidance mechanisms that are less reliant on the quality of detection. With this premise, in this paper, we focus on the range of detection as the guiding measure of the detection quality; or to put in conversely, we are interested in learning avoidance mechanisms that can achieve feasible maneuvers over a wide range of scenarios with the minimum detection range capability.

    A neuroevolution approach based on the AGENT architecture \cite{behjat2019adaptive_arxiv} is utilized here. This approach is hypothesized to be less data hungry (in terms of number of approach scenarios to train) than a supervised learning approach \cite{paul2017bio}. Computing the minimum detection range for any candidate maneuver model (during the learning process) requires solving an internal line search to identify the minimum separation distance condition, over each approach scenario; thus, being able to work with a sparse set of approach scenarios is crucial to a tractable learning process. AGENT or Adaptive Genomic Evolution of NN Topologies is used here to simultaneously work as a classifier and regression model, thereby, producing a heterogeneous output MIMO model. A secondary contribution includes building a classifier to distinguish and discard the state/action pairs where the controller would fail.
    
    
    The remaining portion of this paper is organized as follows: The next section discusses the detection of collision and design of experiments in various uncertain collision scenarios. The following section formulates a framework to avoid collision and generate a trajectory that is followed by the UAVs using a PD controller. The subsequent three sections respectively describe the classifier model training, the evolutionary process of optimizing the neural network topology and weights, and the numerical experiment results. The last section presents the concluding remarks of this paper.

    \section{Reciprocal Collision avoidance}
    \label{sec:framework}

    \subsection{Framework}
\begin{figure}[t]
    		\centering
    		\includegraphics{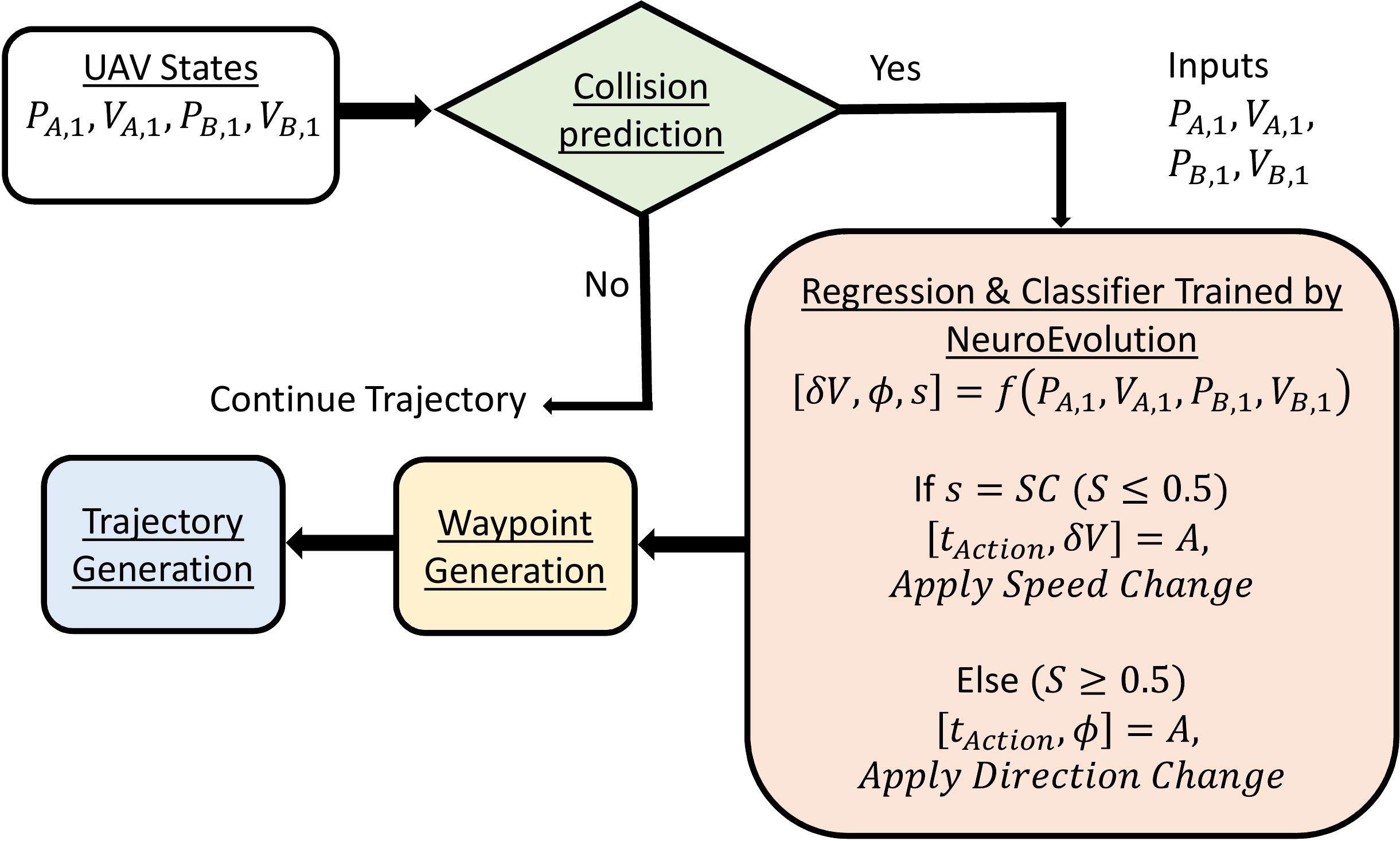}
    		\caption{Collision avoidance framework}
    		\label{fig:flowchart_all}
    \end{figure}

    Figure \ref{fig:flowchart_all} illustrates the collision avoidance framework used in this research. 
    In this method, the UAVs initially detect a potential collision. After detecting the possibility of collision, both UAVs start to decide the necessary maneuver to avoid collision. Two different strategies can be used by the UAV, changing the speed, and deviating the path. The trained neural network finds the best strategy and the parameters associated with that strategy. After taking the optimal decision, the deviation angles and the acceleration are found using surrogate models, and applied to the system. Based on these decisions, a set of waypoints are generated for the maneuver trajectory. A smooth trajectory is then using the minimum snap approach, and a PD controller is employed to follow the trajectory.
    
    \subsection{Collision Detection}
    If the separation between two UAVs becomes lower than a threshold distant $d_{col}$, then it is termed as a collision. The value of $d_{col}$ can be regulated based on the desired level of safety and can be defined in terms of the UAV size (Here, $2\times \text{diameter~of~UAV}$). 
    
    %
    \begin{figure}[t]
      \centering
      
      \includegraphics[width=13cm, height=6.5cm]{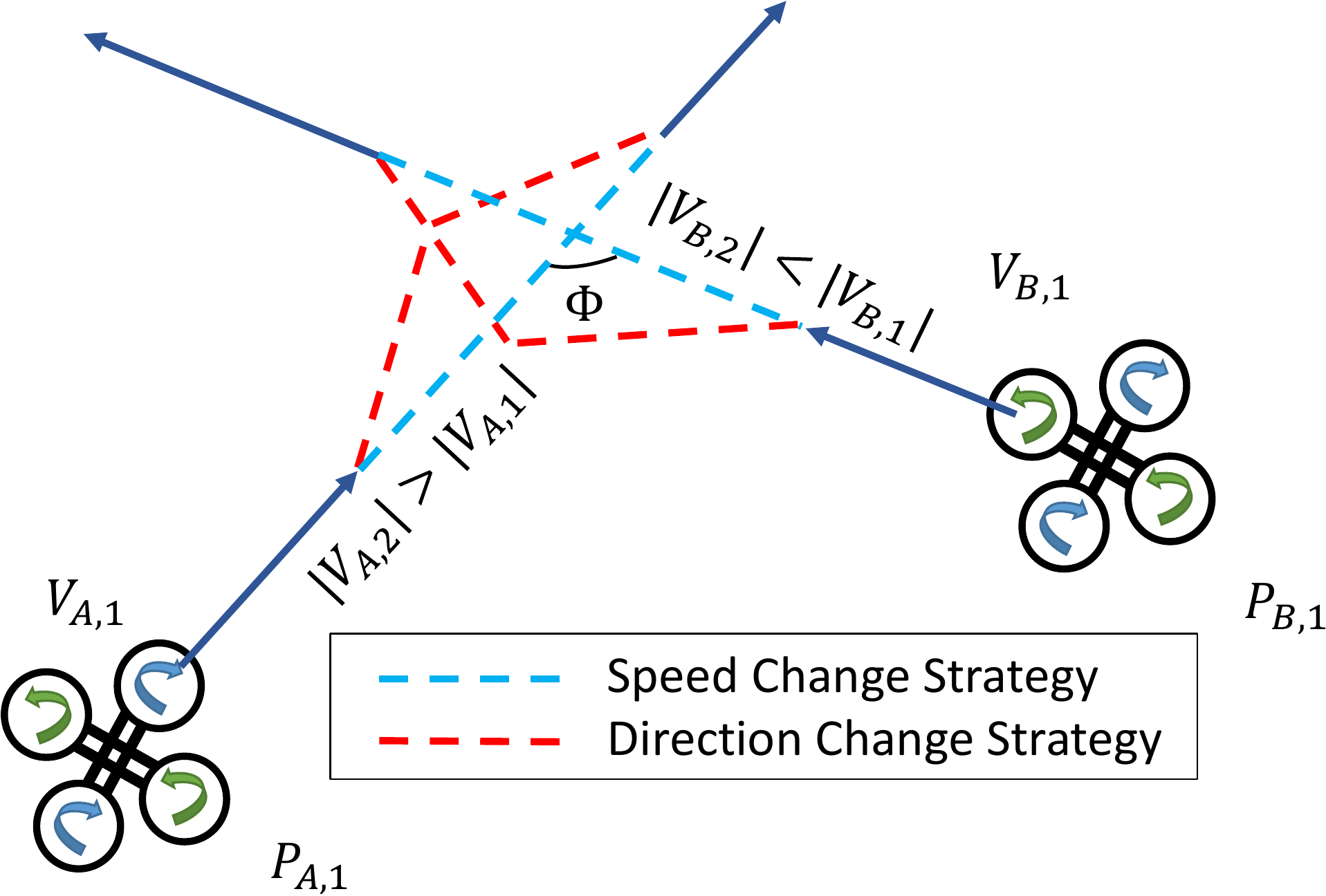}
      \caption{Collision avoidance scenario illustrating the two types of avoidance actions (The UAVs plan and use smooth trajectories guided by these actions)}\vspace{-0.4cm}
      \label{fig:Col}
    \end{figure}
    %
    As shown in Fig. \ref{fig:Col}, let $P_{A,1}$ and $P_{B,1}$ be the current phase of UAVs $A$ and $B$ respectively at a given time point ($t_1$), and $V_{A,1}$ and $V_{B,1}$ are their respective velocity vectors. 
    %
    
    %
    
    The time point at which the two UAVs will come closest to each other can then be estimated as:
    \begin{equation}\label{pt-of-min-sep}
        t_{\min} = \underset{t\in \left[t_1,t_H\right)}
        {{\rm{arg~min}}~ d(t)}
    \end{equation}
    %
    If a solution, $t_H>t_{\min}>t_1$, exists within the time horizon, and $d(t_{min})<d_{col}$, then a collision event is said to have been detected within the time horizon. In that case, the time point of collision is given by:
    \begin{equation}\label{pt-of-collision}
        t_{col} = \underset{t\in\left(t_1,t_{\min}\right)}
        {{\rm{arg~min}}~ t}, ~~{\rm{s.t.}} ~d(t)\le d_{col} 
    \end{equation}
    
    Since the maneuver cannot start before $t_1$, each UAV is then expected to start a collision avoidance maneuver at a designed time, $t_2: t_{1}<t_{2}<t_1+\mu(t_{col}-t_1),~~ 0<\mu\leq1$. The value of $\mu$ can be decided based on a desired safety tolerance.

\begin{figure}[htbp]
    \vspace{-4.7cm}
    \hspace*{-1cm}
      \centering
      \includegraphics[width=23cm, height=15cm]{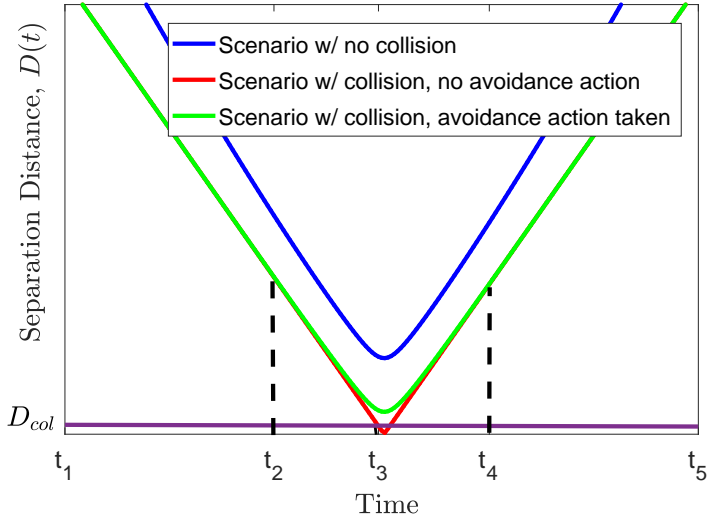}
      \vspace{-2.5cm}
      \caption{Separation distance for illustrative cases with no collision, with collision, and successful collision avoidance. Time points are $t_1$: collision-evasion action decided; $t_2$: evasion maneuver starts; $t_3$: symmetry point of evasion maneuver ($t_3 = t_{col}$); $t_4$: evasion maneuver ends; and $t_5 = t_H$.}
      \label{fig:V_shape}
    \end{figure}

    The variation of the inter-UAV separation for generic collision/avoidance scenarios is illustrated In Fig \ref{fig:V_shape}, where $t_1< t_i < t_H,\ \forall
    ~i=1,\ldots,5$. In this figure, the blue curve shows the separation distance between two UAVs for a scenario where no collision is predicted within the time horizon. The red curve shows the separation distance for a scenario where collision is predicted, and the separation distance curve intersects the collision threshold, $d_{col}$  (yellow line), at time $t_3 = t_{col}$ -- i.e., collision occurs. The green curve shows the separation distance for the scenario where a collision is predicted (originally similar to the red scenario), but an avoidance action is taken to avoid collision successfully. Here `$t_{2}$' and `$t_{4}$' respectively represent the decided time points at which the collision avoidance maneuver begins and ends.
    
    \subsection{Design of Experiments}
    \label{sub_sc_DOE}
    
    A series of collision avoidance scenarios were created in which two identical quadcopter UAVs that are flying in the same horizontal plane have colliding paths. Both UAVs perform collision avoiding maneuvers at the same time. 
    
\begin{figure}[htbp]
      \centering
      \includegraphics[keepaspectratio=true,scale=0.1]{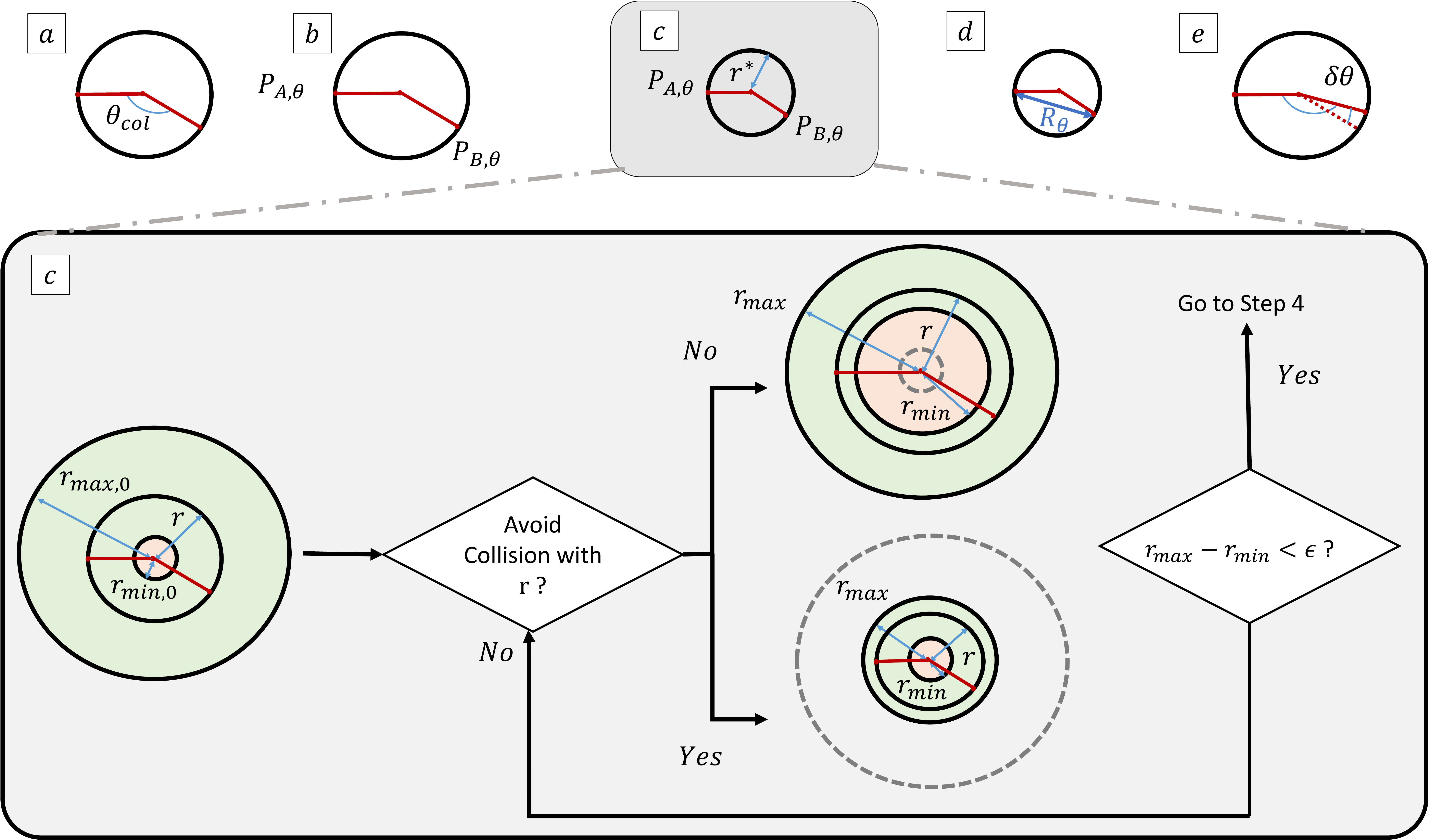}
      \caption{Design of experiment; a) Specify the approaching angle ($\theta_{col}$), b) Find the initial location of UAVs ($P_{A,\theta},P_{B,\theta}$ c) Determine the minimum distance ($r^*$) for specific approaching angle , d) Calculate the minimum detection range for this approaching angle ($R_{\theta}^*$), e) Increase approaching angle by $\delta_{\theta}$}
      \label{fig:DOE_new}
    \end{figure}

    Multiple collision approach angles ($\theta$) are simulated, and a root finding line search is performed over approach scenario to quantify the minimum detection range, $r_{\theta}^*$. Due to computational cost considerations, a frugal set of uniformly spaced 36 different approaching angles ($ \theta_{col} = \{ 5^{\circ}, 10^{\circ} , \cdots 180^{\circ} \}$) were used in training the maneuver model (while a larger set of unseen approach angles/scenarios are used in the test phase).  Figure \ref{fig:DOE_new} illustrates the experiments conducted to estimate the detection range. The Regula Falsi root finding method \cite{mcnamee2013numerical} is used to compute $r_{\theta_i}^*$, which solves the following equation:
    \begin{equation}\label{eq:min_for_regula}
        d_{min} (r_{\theta_i}) = d_{col} 
    \end{equation}
    The variable $r_{\theta_i}$ specifies the collision range and is bounded as $r_{\theta_i} \in [r_{min} ,r_{max}]$. Here $r_{min} = d_{col}$ and $r_{max} =200 m$, reflecting a reasonable range of sensory and controller limitations. With $r_{\theta_i}^*$ computed for each sample scenario $\theta_i$ (where $i=1,2,...,36$), the worst case detection range associated with a given neural network based maneuver model, $f_{ANN}$ (defined in Eq. \ref{eq:neuro_inout}), can be expressed as:
    \begin{equation}\label{eq:min-detection-range}
    \centering
        R_{\min}(f_{ANN}) = \max\limits_{i=1,2,...,36}\left(r_{\theta_i}^*\right)
    \end{equation}
    Based on the current formulation, the upper bound of the objective function will be $2 \times r_{max} $ which is 400m.
    
    
    \subsection{Collision avoidance strategies}
    \label{sub_sc_Collision_avoidance}
    Two different strategies to avoid a collision are utilized in this paper, speed change and direction change, which are further described below.
    
    \textbf{\textit{Speed Change (SC) Strategy}:} In this strategy, two UAVs accelerate and decelerate with the same amount but with the reverse order. While the faster UAV initially decelerates  from $t_{2}$ to $t_{3} = t_{col}$, the slower UAV initially accelerates. After reaching the collision point they follow reverse actions (faster UAV accelerates $t_{3}$ to $t_{4}$, and slower UAV decelerates during this time period) in a way that at $t_5$ they are in their original desired position and speed as illustrated in figure \ref{fig:Col}. 
    The  SC strategy is defined in terms of the average change in speed, $\delta_V$, and the time point, $t_2$ when the SC maneuver initiates. The time point is the same for both UAVs, and the magnitude of acceleration and deceleration is also equal. 
    The average velocity of the UAVs during the avoidance maneuver, i.e., between time points $t_2$ to $t_3$ and $t_3$ to $t_4$ are given by:
    \begin{equation}\label{SC-stage-1-vel}
        \begin{split}
            & \overline{V}_{A,2-3} = {(|V_{A,1}| - \delta_V)}.\frac{V_{A,1}}{|V_{A,1}|}\\
            & \overline{V}_{B,2-3} = {(|V_{B,1}| + \delta_V)}.\frac{V_{B,1}}{|V_{B,1}|}\\
        \end{split}
    \end{equation}
    \begin{equation}\label{SC-stage-2-vel}
        \begin{split}
            & \overline{V}_{A,3-4} = \frac{V_{A,1}(t_{4} - t_{2}) - V_{A,2-3}(t_{3} - t_{2})}{t_{4} - t_{3}}\\
            & \overline{V}_{B,3-4} = \frac{V_{B,1}(t_{4} - t_{2}) - V_{B,2-3}(t_{3} - t_{2})}{t_{4} - t_{3}}
        \end{split}
    \end{equation}
    The maximum time to initiate the strategy is 3 seconds, and the maximum change of speed is 5 m/s.
    
\begin{figure}[t]
      \centering
      \includegraphics[width=13cm,height=7cm]{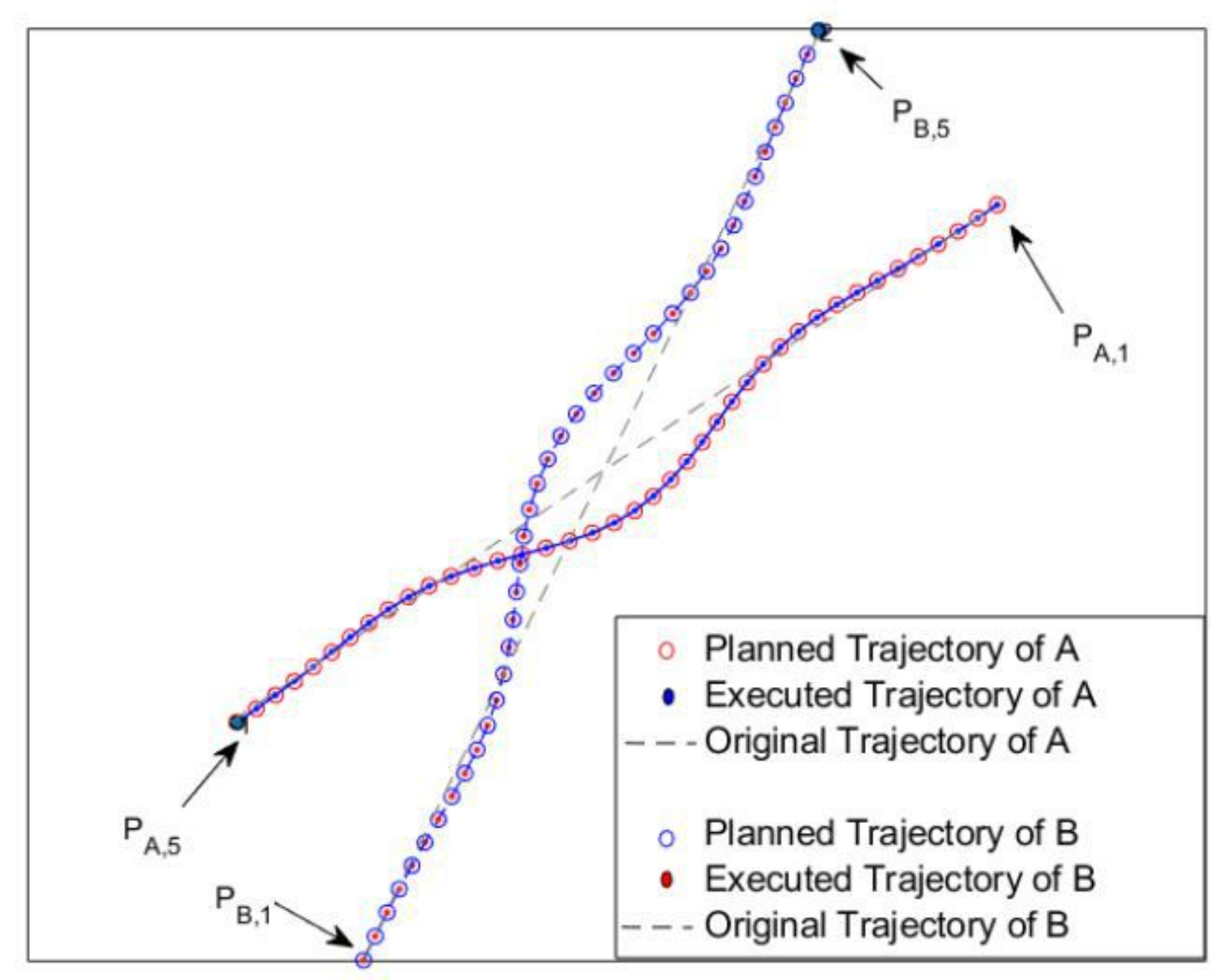}
      \caption{Direction change (DC) strategy: illustrating planned and executed trajectories under this strategy}
      \label{fig:Direction_change_path}
    \end{figure}

    \textit{\textbf{Direction Change (DC) Strategy:}} In this strategy, both UAVs deviate from their original path, and then they return to their original path. In order to have a mutually coherent trajectory planning for UAVs, we make both UAVs deviate to their left side (Counter clockwise turn) at  $t_2$. At time point $t_3 = t_{col}$, UAVs start to return to their original path and they plan to be in their original path at $t_4$ (clockwise turn). A smooth trajectory is planned to execute the maneuver and UAVs plan to return to their original desired position and velocity at $t_5$. 
    Figure \ref{fig:Direction_change_path} provides a representative illustration of a DC maneuver, showing both the planned and the controller executed trajectory of the UAVs.
    
    Both strategies are designed in a way that the maneuvers do not delay (or advance) entire operation after these maneuvers are finished which is vital for sensitive missions like \cite{oil} where entire operation schedule does not have flexibility.
    
    
    Because the deviated path is longer than the original path,  both UAVs need to increase their average speed to preserve their pre-planned (time encoded) mission paths following the maneuver. Their average velocity during the DC maneuver can be expressed as:
    \begin{equation}\label{DC-stage-1-vel}
       \begin{split}
          \overline{V}_{A,2-3} = \frac{1}{\cos\phi} R\ V_{A,1} \\
          \overline{V}_{B,2-3} = \frac{1}{\cos\phi} R\ V_{B,1} 
       \end{split}
    \end{equation}
    where $R$ is the standard rotation matrix in 2D space. The maximum allowed time to initiate the strategy (from the point of collision detection) is 3 seconds, and the maximum change of speed is $30^{\circ}$. 
    %
    
    \subsection{Maneuver Planning System}
    
\begin{figure}[!ht]
      \centering
    	  \includegraphics[width=15cm, height=10cm]{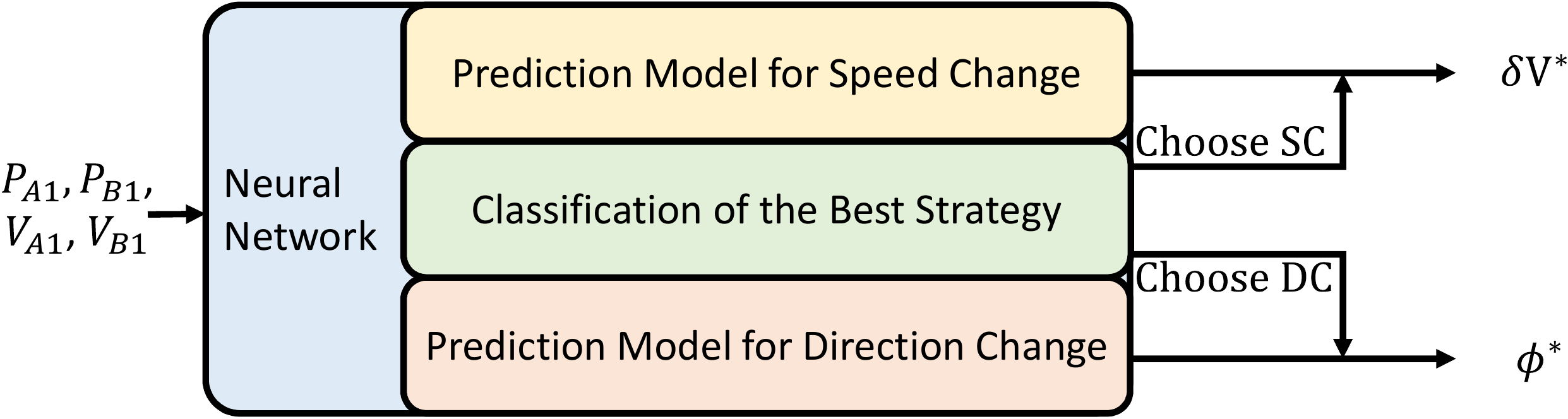}
      \caption{Online maneuver planning module in operation}
      \label{fig:predictionmodel_working}
    \end{figure}

    For each input, the classifier portion of the trained neural network decides the better strategy among speed change (SC) and direction change (DC), and regressor portion determines the parameters encoding the strategy. The inputs to the neural network model are a subset of the state parameters defining the initial position and velocities of the UAVs. Figure \ref{fig:predictionmodel_working} illustrates the strategy classification and  decision-making module.
    
    \subsection {Trajectory Planning}
    
    The minimum snap trajectory planning approach \cite{minimumsnap} is used here. This method uses a seventh order polynomial (spline) to fit the sequence of setpoints or waypoints estimated from the parameters of maneuver decided by the neural network (e.g., deviation angle in DC). These polynomials produce the time stamped position command, which is to be followed by the controller.
    
    

    \section{MULTI FIDELITY MODELING }
    \label{sc:surr_model}
    
    The proposed training approach needs a simulation that is relatively inexpensive to run. To this end, the minimum distance of separation is calculated using a geometrical model instead of using the simulation.  
    
    \subsection{Minimum Distance Model}    
    The minimum distance between two UAVs can be calculated using the simulation model, but running the dynamics simulation is expensive (in the context of the large number of function evaluations needed to train). In order to decrease the computational cost, the trajectory points are used to estimate the minimum distance between the UAVs. 
    Equation \ref{eq:min_dis_by_traj} explains this process
    
    \begin{equation}
        \label{eq:min_dis_by_traj}
        \begin{aligned}
        &P_{A}(t) = \sum_{i=1}^{4} q_{A,i}(t) H(t-t_i)H(t_{i+1}-t)\\
        &P_{B}(t) = \sum_{i=1}^{4} q_{B,i}(t) H(t-t_i)H(t_{i+1}-t)\\
        &D(t) =  |P_{A}(t) -P_{B}(t)|
        \end{aligned}
    \end{equation}
    
    Here $q_{A,i},q_{B,i}$ are the $7^th$ order polynomials between each consecutive waypoints in the path and $H(t)$ is the unit step function.The minimum point is estimated by computing the roots of $\frac{\partial D(t)}{\partial t}$.
    
    \subsection{Classification of Actions based on Controller Failure}    
    Using the trajectory to find the minimum distance decreases the computational cost because generating trajectories is less expensive compared to simulating the dynamics. However, it can cause an error if the controller is unable to keep the UAV in its desired trajectory or causes loss of stability. Therefore it is essential to identify and have a mechanism to  discard situations where the controller would fail in the above-stated manners. To this end, a classifier is trained over a set of trajectory samples encapsulating the range of trajectories planned for collision avoidance maneuver. 
    The properties of the trained classifier are summarized in table \ref{tab:Classifier}.
    
    \begin{table}[htbp]
    \caption{Parameters of classifier for controllable decision} 
    \begin{center}
    \label{tab:Classifier}
       \begin{tabular}{l c c} 
          \hline \vspace{0.1cm}
         \textbf{Parameter} &
        \textbf{Direction Change } &
        \textbf{Speed Change} \\
        \hline \vspace{0.1cm}
         Input &   $V_{1}$, $\phi$ &   $V_{1}$,  $\delta_V$\\  
    
         Total Samples &  1000 & 1000\\
         Training Algorithm & SVM (Cubic)  & Bagged Trees \\
         Cross Validation & 10-fold & 10-fold\\
         Misclassification  & $13.2\%$ & $13.9\%$  \\
         \hline
       \end{tabular}
    \end{center}
    \end{table}

    \section{NEUROEVOLUTION: \textit{TRAINING AVOIDANCE MANEUVER}}
    \label{sc_Optimization}    
        \subsection{Optimization Problem Definition}
        \label{sub_sc_Desision}
        
        
    Based on our observations in \cite{behjat2019adaptive_arxiv}, using smaller reaction time leads to a more controllable maneuver; these maneuvers are not necessarily energy optimal. Since the focus of the current paper is not about energy optimality but rather the detection quality which might be more useful in practice, it is possible to fix the reaction time to the smallest practical value. A value of $0.1s$ is used in the current study. The input to the neural network based maneuver model includes the initial state parameters of the UAVs, and the output will be the strategy choice (Sc vs. DC) and its parameters (change in speed or change in heading angle). Equation \ref{eq:neuro_inout} summarizes the designed input and the output of the maneuver model.
        
        
        \begin{equation}
                \label{eq:neuro_inout}
                 [\delta_V, \phi , s] = f_{ANN}(P_{A,1},P_{B,1},V_{A,1},V_{B,1})
        \end{equation}
    Where $\Delta P_{X,1}$ is  $ P_{A,X,1} - P_{B,X,1}$. Also $\Delta V_{X,1},\Delta V_{X,1}$  are the initial velocity in  X and Y directions.
       
    In equation \ref{eq:neuro_inout}, $f_{ANN}$ indicates the neural network model.  In this equation, each term is a $2$ element vector; a total of $8$ parameters. Based on the current DoE, it is possible to reduce the input space to four independent variables. 
    The output $s$ is used as a classifier; if $s \leq 0.5$ then speed change strategy is used, otherwise the direction change strategy is used. 
    

    The objective function, is defined as minimization of required detection range (as explained in section \ref{sec:framework} \ref{sub_sc_DOE}), i.e., minimize $f=R^*$.
        
    
    
    Table \ref{tab:para_opt}  lists the UAV specifications and parameters used for training the maneuver model.

    \begin{table}[htbp]
     \begin{center}
     \caption{UAV Specifications}
         \begin{tabular}{l c} \hline 
             \textbf{Parameters} &\textbf{Value} \\
             \hline 
             Quadcopter UAV structure & X Shape \\
             UAV Weight & 1 kg\\
             UAV Initial speed & 16.67 m/s \\
             UAV Max Thrust & $8 \times$Weight\\
             UAV size (diameter) & 0.71 m  \\
            Safe separation distance ($d_{col}$) & $2 \times$ diameter \\
             Reaction time ($t_{col}$) & 0.1 seconds\\ 
             \hline
           \end{tabular}
        \label{tab:para_opt}
     \end{center}
    \end{table}
    
    

    \subsection{Neuroevolution Process}
    \label{sub_sc_Neuro}
    
    The optimization method cannot use the labeled data because the optimal answer needs asymmetric decision of the second UAV. Therefore instead of supervised learning, we follow a neuroevolution method \cite{floreano2008neuroevolution} to train the neural network. More specifically, we exploit our own recently developed neuroevolution method that simultaneously optimizes the topology and weights of the neural network, and uses diversity and fitness balance controllers to prevent premature stagnation and robust convergence \cite{stanley2002evolving,vargas2017spectrum}. Further description of our neuroevolution method called AGENT can be found in \cite{behjat2019adaptive_arxiv}. Neuroevolution uses an evolutionary algorithm to optimize the topology and weights of a neural network, and is typically used for solving problems that can be posed as reinforcement learning \cite{salimans2017evolution}; however unlike RL, neuroevolution is significantly more amenable to parallel deployment and escaping local minima, both crucial to expensive offline learning investments.
    
    
    
    Here, the process of training the neural network via neuroevolution can be perceived as minimizing a cost function that is given by Eq. \ref{eq:min-detection-range}. Although, explicitly a single objective unconstrained optimization is being solved here during neuroevolution, consideration of two constraints are implicit to the computation of Eq. \ref{eq:min-detection-range}, namely ensuring that the trajectory is collision free and flyable by the controller.

       \section{RESULTS AND DISCUSSION}
    \label{sc_results}
    The AGENT framework is used to train the neural network maneuver model using a population of 100 and allowed maximum number of generation set at 30. The neuroevolution process was executed on a Intel Xeon CPU ES-1620 v4 16GB RAM workstation, and completed in only about 3.5 hours -- which is remarkable for RL-type problems involving systems as complex as quadcopter UAVs. This attractive computational efficiency is partly attributed to the robust exclusion of the dynamics/controls simulations (by using the geometric trajectory model and classifier discarding actions where controller would fail). 
    \subsection{Performance Analysis of the Trained Maneuver Model} 
        \begin{figure}[t]
    \begin{subfigure}{.5\textwidth}
      \centering
      \includegraphics[keepaspectratio=true,scale=0.1]{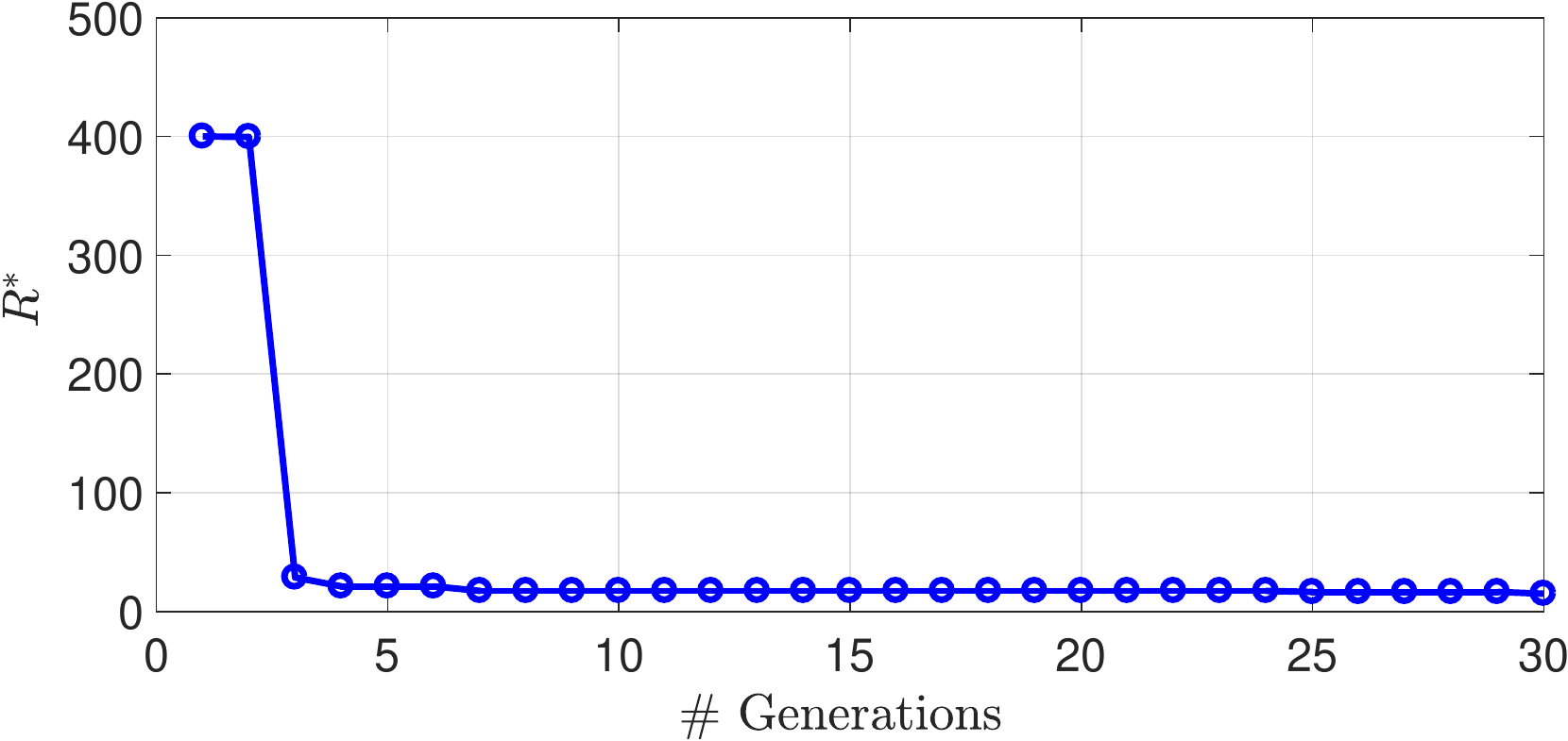}
      \vspace{+.5cm}
      \caption{Convergence history for AGENT}
      \label{fig:conv_agent}
    \end{subfigure}%
    \begin{subfigure}{.5\textwidth}
      \centering
      \includegraphics[keepaspectratio=true,scale=0.05]{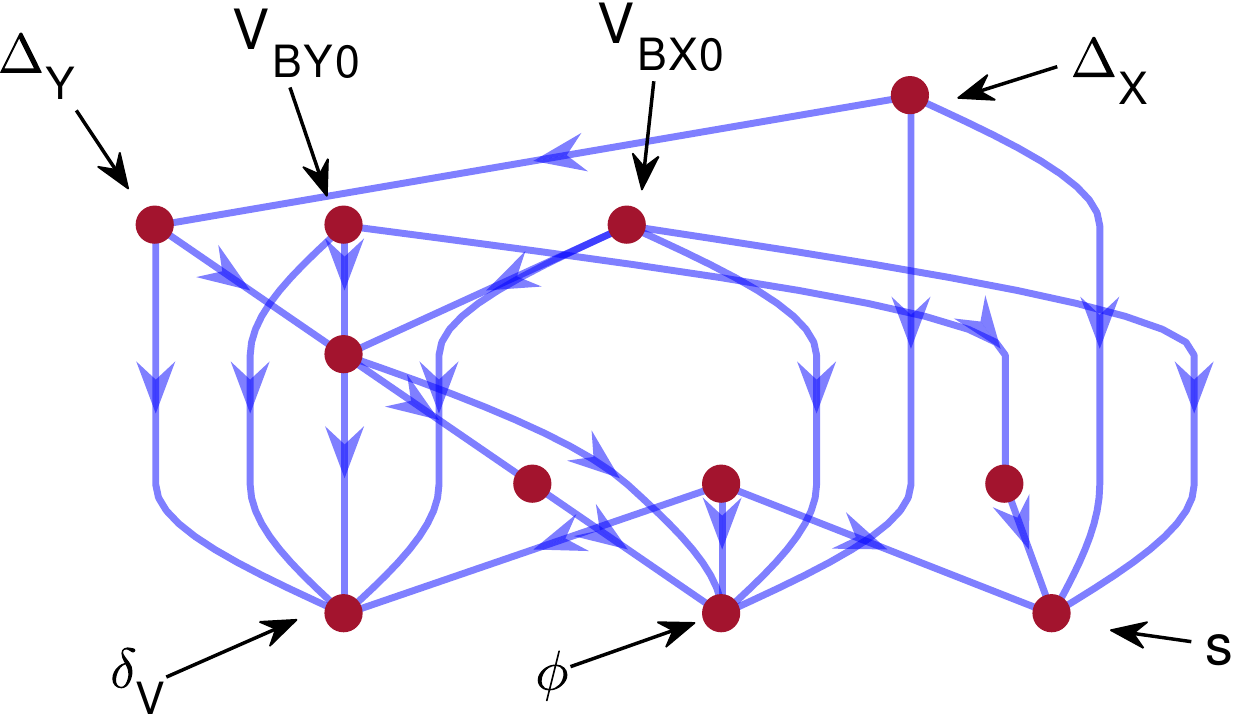}
      \caption{Optimized ANN topology, comprising 4 hidden neurons}
      \label{fig:final_net}
    \end{subfigure}
    \caption{Neuroevolution progress and outcome: training maneuver model}
    \end{figure}
        
    Figure \ref{fig:conv_agent} and \ref{fig:final_net} respectively show the convergence history of AGENT and the structure of the optimized neural network given by AGENT. The first feasible maneuver model is found in generation 3, and gradually optimized further over the next $\sim$27 generations.

\begin{figure}[tbp]
      \centering
       \includegraphics{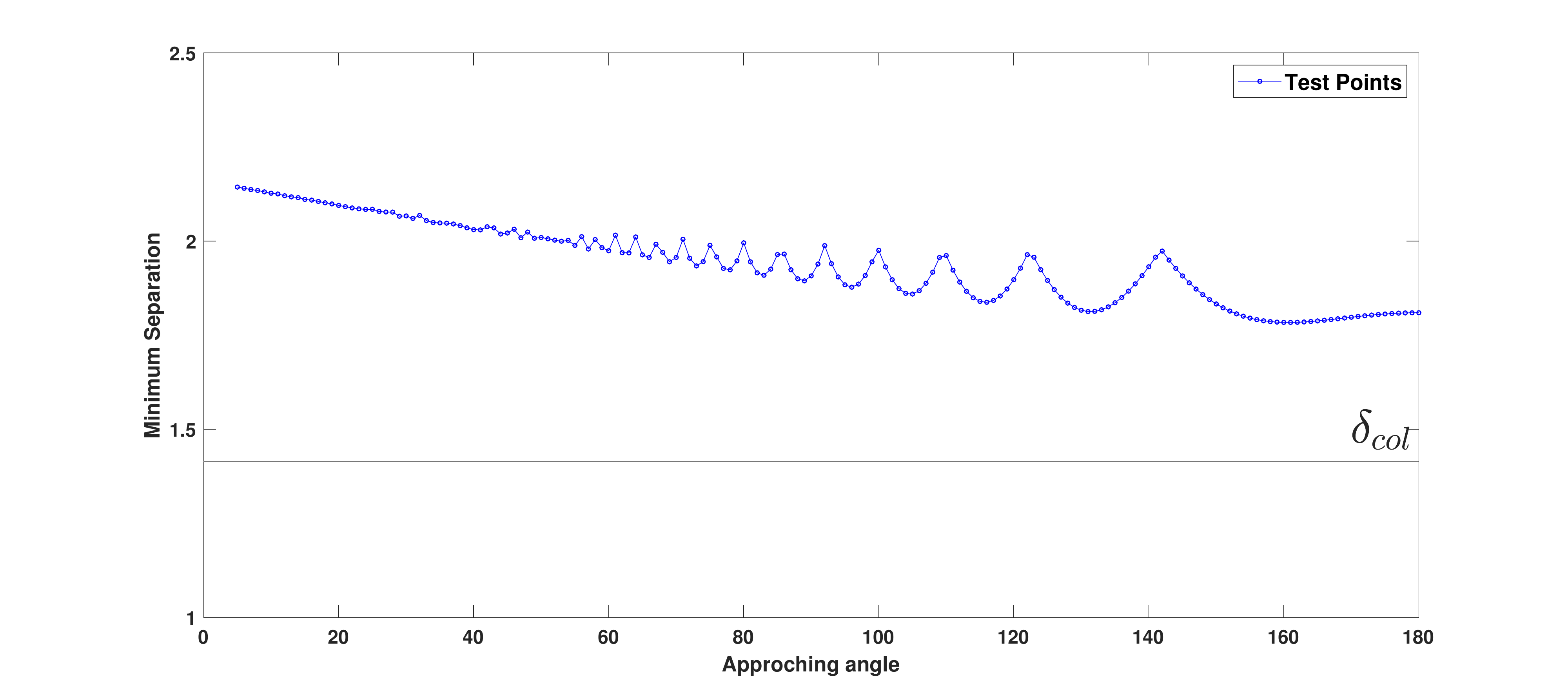}
      \caption{Minimum separation distance resulting from maneuver executed by AGENT-derive model, over all test scenarios}
      \label{fig:all_tests}
        \end{figure}

    In order to evaluate the generalizability of the model produced by AGENT, it was tested on a suite of 180 scenarios defined by $ \{ 1^{\circ}, 2^{\circ} , \cdots 180^{\circ}  \}$. Note that, neuroevolution trained the neural network (maneuver) model over a small set of scenarios that comprises 20\% of the test samples. Figure \ref{fig:all_tests} shows the minimum separation of UAVs under the maneuver decided by the trained ANN model. The minimum separation distance decreases (with oscillations, which might be an artifact of the sparse sample trained ANN model), as the approaching angle increases, while safely staying above the collision threshold, $\delta_{col}$.
    
     \begin{figure}[htbp]
    \begin{subfigure}{.5\textwidth}
      \centering
      \includegraphics[width=9cm, height=6.5cm]{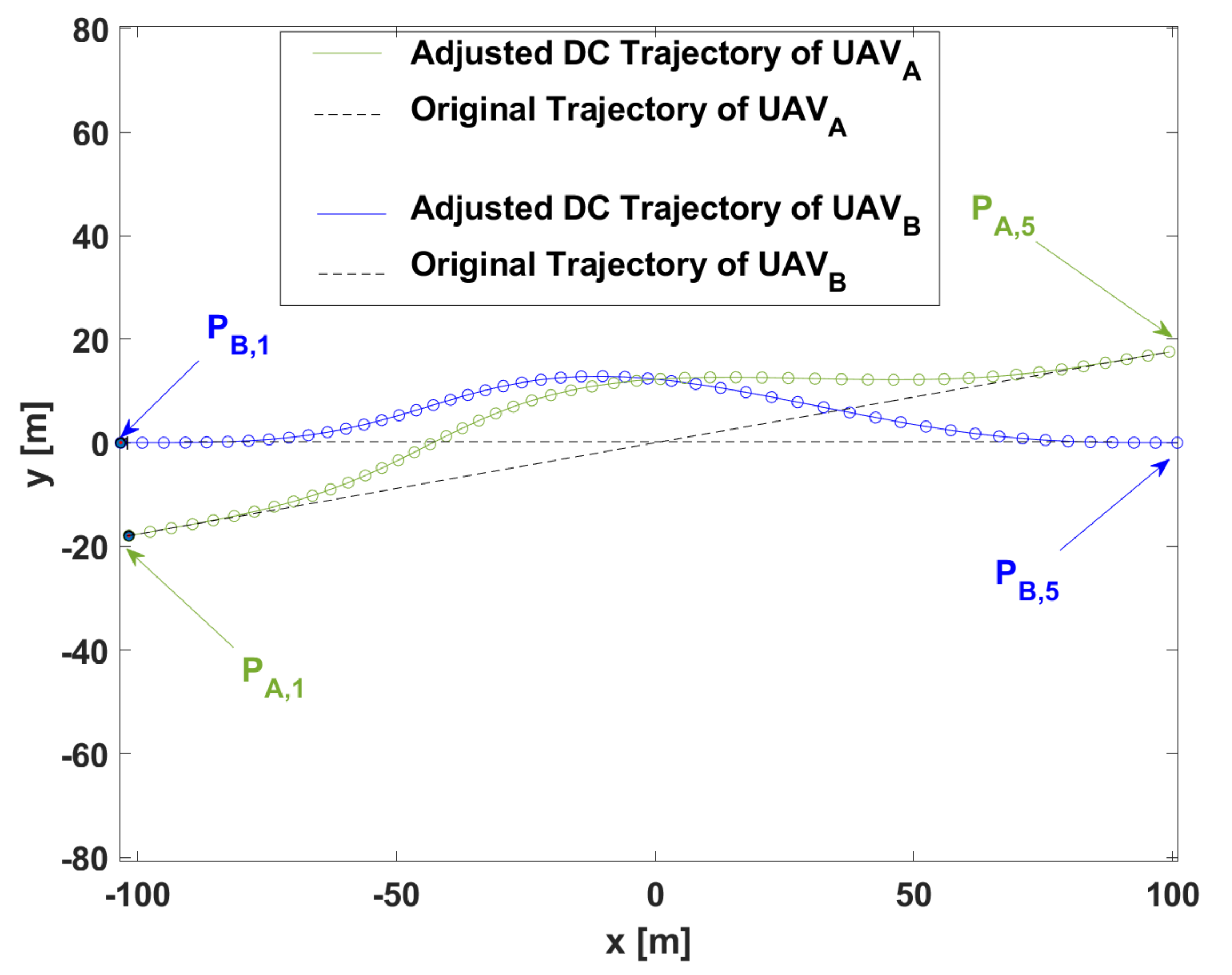}
      \caption{Acute angle ($10^{\circ}$) collision avoidance maneuver}
      \label{fig:10a}
    \end{subfigure}%
    \begin{subfigure}{.5\textwidth}
      \centering
      \includegraphics[width=9cm, height=6.5cm]{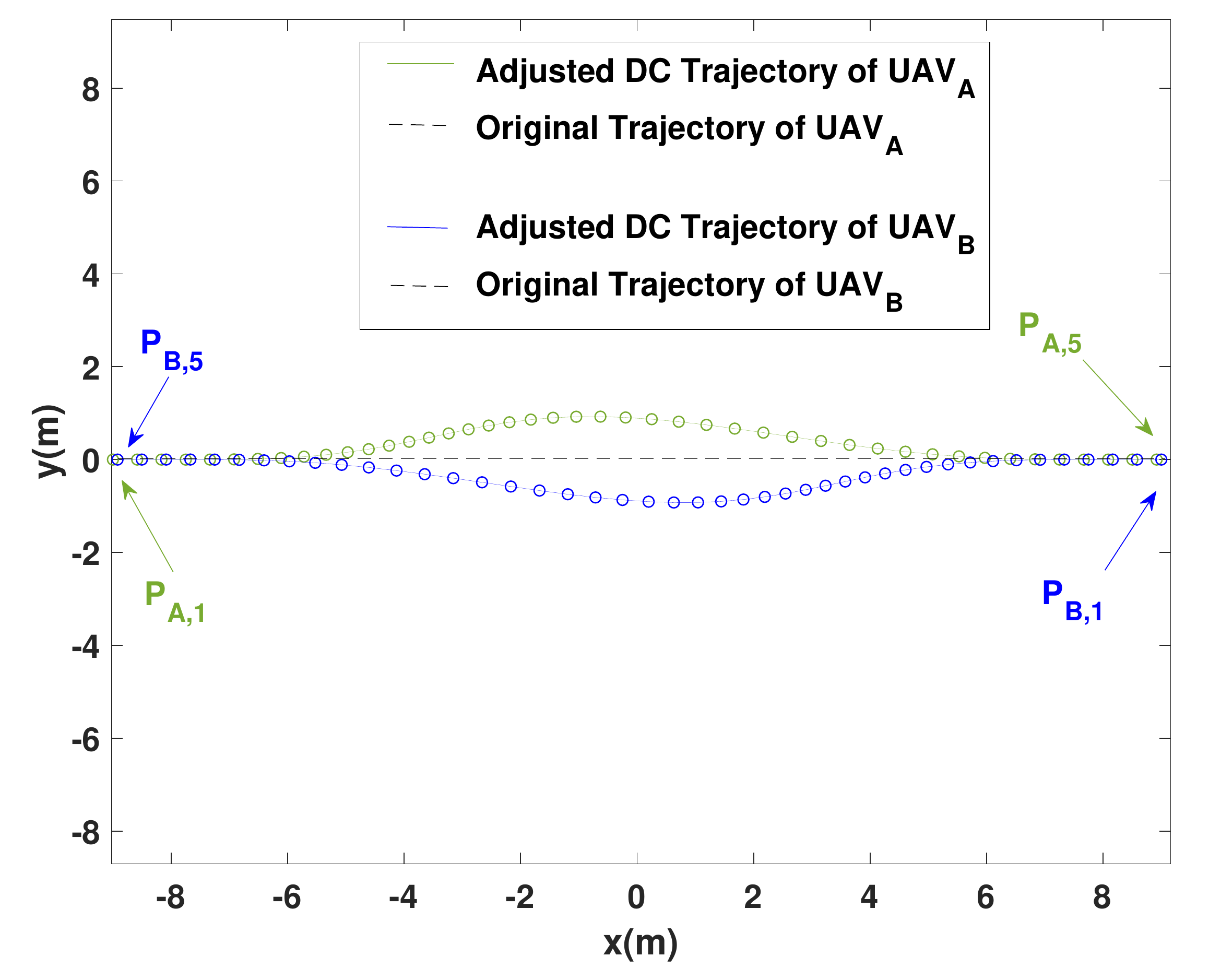}
      \caption{Head-on ($180^{\circ}$) collision avoidance maneuver}
      \label{fig:10b}
    \end{subfigure}
    \caption{Modified trajectories under maneuver for different approach scenarios (computed using high fidelity control/dynamics simulations for validation purposes)}
        \end{figure}

    In addition, we test the performance of the trained maneuver model with the high fidelity simulations that include solving the full control/dynamics. Figures \ref{fig:10a} and \ref{fig:10b} show the original and modified paths under the maneuver respectively for an acute approaching angle ($10^{\circ}$) and head-on collision ($180^{\circ}$). The sustainability of the performance is evident from these figures. 
    These results also show that the maneuvers for acute angle collisions lead to a longer path, which might be less energy efficient, compared to those needed for head-on collisions.


    \subsection{Comparison with Offline Optimized Maneuver} 
\begin{figure}[htbp]
      \centering
      \includegraphics{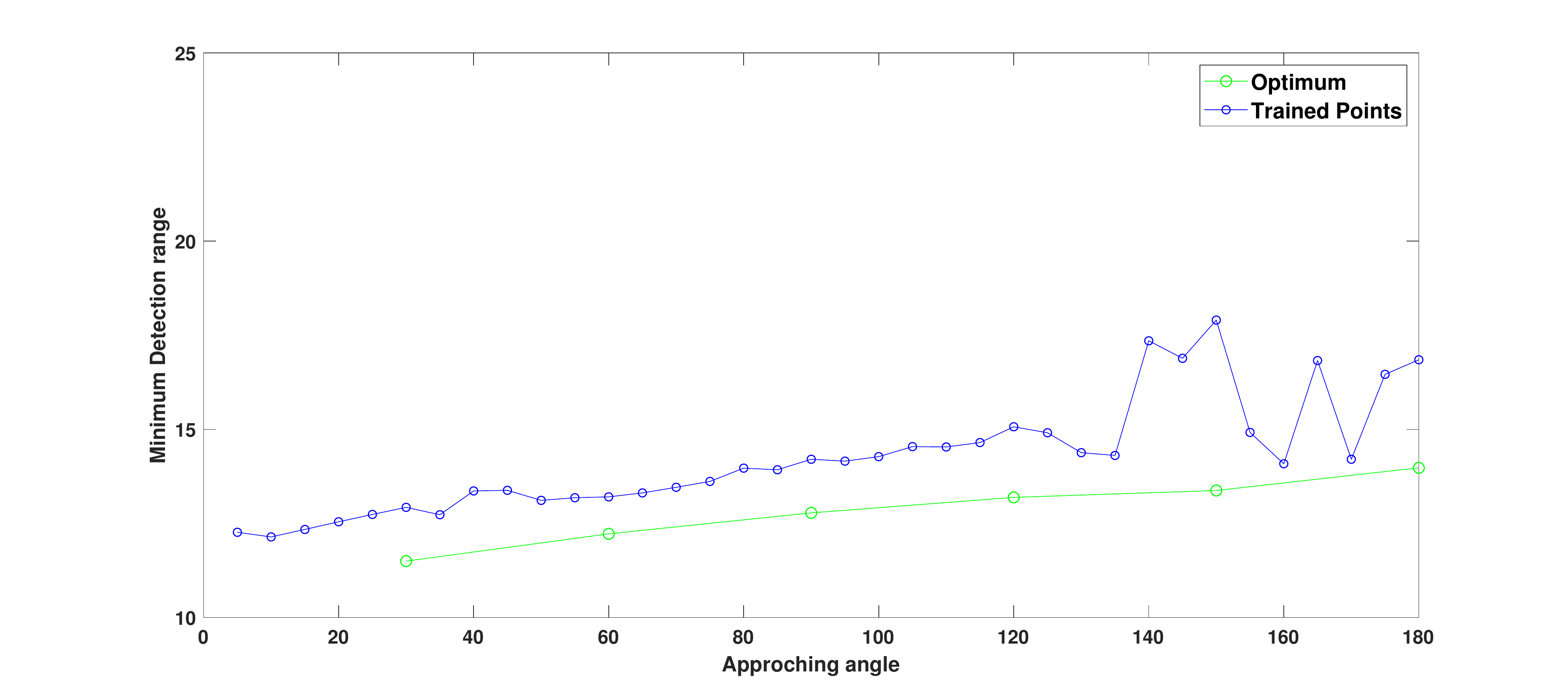}
      \caption{Comparison of minimum detection range given by model obtained via neuroevolution and that given by offline (PSO-based) optimization over individual approach scenarios}
      \label{fig:min_req}
        \end{figure}

    In order to assess the closeness to optimality of the decisions made by the neural network maneuver model, its associated minimum detection range was compared with that given by a global offline optimization process (operating on individual scenarios). This comparison is done for a small variety of six approach angles. For this purpose, the minimum detection range was obtained offline using a Particle Swarm Optimization (PSO) algorithm \cite{MDPSO}, where a population of 60 and maximum allowed iterations of 20 is used. PSO is chosen here based on prior work; where different gradient based and gradient-free algorithms were tested \cite{paul2017bio}. To put this comparison into the cost perspective, if the total investment of sampled scenarios is kept fixed at that allowed for neuroevolution, then a supervised learning approach using PSO-optimized outcomes would only get a challengingly frugal set of 90 samples to train three (one classifier and two regression) models on.   
    Figure \ref{fig:min_req} shows the comparison of the resulting minimum detection range obtained by the AGENT-trained model and that given by PSO. While, the offline (PSO) optimization expectedly finds a better solution for the six tested scenarios compared to the generalized model trained by neuroevolution, the difference is small ($<2$m), except for approach angles greater than 140$^\circ$.
    The latter is a topic of future exploration, especially considering the oscillatory behavior of AGENT-derived model observed here and in Fig. \ref{fig:all_tests}, w.r.t. the higher approach angles. It is however important to note that both the AGENT-derived ANN-based maneuver model and the PSO outcomes show an overall similar trend, where the minimum required detection range increases with increasing approach angles. Intuitively, this could be related to the greater relative velocity of approach (which closes the distance between UAVs faster, demanding larger detection range) in the case of ``close to head-on" approach scenarios, when compared to smaller approach angle scenarios.

    \section{CONCLUSION}
    
    \label{sc_conclusion}
    This paper presents a new approach to online planning of collision avoidance maneuvers in the case of cooperative UAV-UAV encounters. The planning objective focused on minimizing the required detection range, mainly to explore the reliance on sophisticated sense/detection approaches and their perfect operation (often challenging in practice), and in turn demonstrate a potential solution to reducing this dependency. In this context, we build upon our work on reciprocal collision avoidance for identical multi-rotor UAVs. Artificial neural networks (ANN) are used to serve as the online maneuver model, which, given the initial state of the own and peer UAV, simultaneously outputs the maneuver strategy to choose (direction change vs. speed change) and their parameter values (e.g., amount of change in heading). To enable the construction of a generalizable ANN-based maneuver model while using a small set of sample collision scenarios, we exploit the topology-varying neuroevolution paradigm (specifically, our own variation called AGENT) that seeks to solve RL-type problems. Other approximations are made in the form of using a geometric model for computing the minimum separation distance between UAVs and a pre-trained classifier to discard state/action pairs that cannot be reliably flown by our designed PD controller -- together, they alleviate the need for using the complete dynamics/controls simulation during the neuroevolution process. Together, these approaches allowed an attractive training time scale, amenable to desktop workstation-based learning. Successful validation was performed using high fidelity dynamics simulations to show the effectiveness of the maneuvers generated by the trained model, even though the training used the various efficiency-facilitating approximations.
    
    The maneuver model resulting from neuroevolution performed successfully (in terms of avoiding collisions) over the training and unseen test scenarios, spanning different approach angles. Some oscillatory behavior was observed w.r.t. higher approach angles (closer to head-on scenarios), which also caused greater deviations from offline-optimized results. Future work would focus on investigating this oscillatory behavior issue, w.r.t. sample size and learning settings used for training. In addition, the current work has focused mainly on a sort of niche domain of identical UAVs operating on the same horizontal plane, and executing maneuvers that preserve the flight altitude; to analyze the flexibility of our collision avoidance maneuver learning framework, future extensions to non-identical UAVs and altitude changing maneuvers is needed. Together with testing on physical platforms, these further research directions would help open up opportunities of using such neuroevolutionary learning mechanisms to design autonomous behavior of UAVs. 
    
    

    \section*{Acknowledgement}
    Support from the DARPA Award HR00111890037 from Physics of AI (PAI) program is gratefully acknowledged.
    
    \bibliographystyle{aiaa}

    \clearpage    
    
    \end{document}